\documentclass[11pt]{article}

\usepackage[preprint]{acl}

\usepackage{times}
\usepackage{latexsym}

\usepackage[T1]{fontenc}

\usepackage[utf8]{inputenc}

\usepackage{microtype}

\usepackage{inconsolata}

\usepackage{graphicx}

\usepackage{amsmath}
\usepackage{multirow}
\usepackage{makecell}

\title{Top-down string-to-dependency Neural Machine Translation}

\author{Shuhei Kondo \\
  Unaffiliated \\
  \texttt{shuhei-k@proton.me} \\\And
  Katsuhito Sudoh \\
  RIKEN Center for Advanced Intelligence Project \\
  Nara Women's University \\
  \AND
  Yuji Matsumoto \\
  RIKEN Center for Advanced Intelligence Project \\
  \\}

\begin{document}
\maketitle
\begin{abstract}
Most of modern neural machine translation (NMT) models are based on an encoder-decoder framework with an attention mechanism. While they perform well on standard datasets, they can have trouble in translation of long inputs that are rare or unseen during training. Incorporating target syntax is one approach to dealing with such length-related problems. We propose a novel syntactic decoder that generates a target-language dependency tree in a top-down, left-to-right order. Experiments show that the proposed top-down string-to-tree decoding generalizes better than conventional sequence-to-sequence decoding in translating long inputs that are not observed in the training data.
\end{abstract}

\section{Introduction}
Sequence-to-sequence (S2S) models with an attention mechanism has become a de facto standard of neural machine translation (NMT). 
Early S2S models are based on RNNs \cite{bcb15, luong}, and the more recent Transformer model \cite{transformer}, primarily based on self-attention, has replaced them.

In a typical S2S model, an encoder takes a source language sequence as its input and a decoder generates its translation sequentially, referring to source representations through the attention mechanism. 
Positional information and history are handled by recurrence of hidden states in RNN-based models and self-attention in the Transformer model.
These can cause trouble in translation of long sentences, because the former must store all the history in a fixed-sized set of hidden states and the latter relies on positional embeddings that might be unseen during training.

There is another factor which can make translation of long sentences more challenging.
It is a standard practice in NMT training to exclude long sentences from the training data since the era when statistical approaches were the mainstream.
This strategy is also adopted in NMT training, mainly for computational efficiency.
While such technique is prevalent, its implication for translation of long sentences has been largely overlooked. 

In the field of language modeling, one approach to model such long-distance dependencies is to incorporate syntax information \cite{kuncoro-etal-2018-lstms}.
While some studies suggest that NMT encoders can capture source syntax implicitly \cite{lstm-syntax, tf-syntax}, S2S models do not use such information explicitly. 
Recently, there have been several studies that aim at incorporating syntax explicitly into NMT.
Some models focus on source syntax either by adding structural labels to sequential encoders \cite{modeling} or with tree-based encoders \cite{t2s, t2t}.
Other models incorporate target syntax into decoders, either by augmenting the decoder outputs with syntactic information \cite{ag-s2t, an, s2ccg, multi} or designing tree-based decoders \cite{msra-s2dep, trdec}.

As our main concern is to improve translation of long sentences, we focus on models that use syntactic information in the target language. 
Since we can expect the input sequence to be a valid sentence and the encoder to capture source syntax implicitly, a decoder with target syntax is more important and beneficial to deal with syntactic phenomena such as long-distance dependencies and to encourage syntactically plausible outputs.

We propose a novel syntactic decoder that generates target-side dependency trees recursively in a top-down, left-to-right order.
While our decoder uses a single stacked LSTM that generates dependency structures and surface words with shared parameters and hidden states, our model differs from S2S models with syntax-augmented outputs in that it maintains a stack of actions necessary to complete the generation process and uses it to inform the decoder.

This paper's main contributions are two-fold.
\begin{enumerate}
  \item We show that the translation quality of conventional S2S models (including the Transformer) starts to drop rapidly as inputs get longer beyond the threshold which they are trained with, and standard test sets are not enough to capture this phenomenon because they contain only a limited amount of long inputs.
  \item Our experiments with three training datasets with different length threshold and a non-standard test sets extracted randomly from the Tilde MODEL corpus \cite{tilde} show that the proposed top-down model generalizes better than plain S2S models to long sentences that are not observed during the training process.
\end{enumerate}
We will release the code on \url{https://github.com/shuheik/topdown_nmt}.

\section{Related Work}
\subsection{S2S models with Target Syntax Information}
Some studies incorporated target-side syntax into S2S architectures by generating target sequences augmented with linguistic information instead of plain (sub)word sequences.
\newcite{ag-s2t} proposed an S2S model that predicts a linearized, lexicalized constituency tree on the target side.
\newcite{an} proposed a similar model with dependency structure.
\newcite{s2ccg} introduced the Combinatory Categorial Grammar (CCG) into an S2S model by predicting subwords and CCG supertags alternately.
In addition to linearized trees, \newcite{multi} proposed two new target representations for S2S models. 
One is a linearized derivation and the other is an interleaved sequence of part-of-speech (PoS) tags and words. 
They show that an ensemble of models with different output representations leads to better results.

\subsection{Models with Syntactic Decoders}
\newcite{msra-s2dep} proposed a sequence-to-dependency NMT model in which the decoder consists of a word-RNN and an action-RNN that run jointly to predict target words and their dependency structure. 
Essentially, the action-RNN is an arc-standard shift-reduce parser \cite{arc-standard} and the word-RNN generates a target word when a Shift action is predicted by the action-RNN.
The target syntactic context is computed from the parser configuration with a feed-forward neural network and used for word prediction.

\newcite{trdec} proposed a tree-based decoder that contains a rule-RNN and a word-RNN.
In their model, the rule-RNN predicts target-side production rules in depth-first, left-to-right order and the word-RNN generates a subword sequence that corresponds to a preterminal in the production rules.
Note that the rule-RNN predicts production rules, which is a sequence of syntactic labels such as non-terminals.
As a consequence, the decoder cannot predict production rules that are unseen in the training data. 
They compare several target tree structures and obtain the best BLEU score with non-syntactic balanced binary trees.

\newcite{li-etal-2023-explicit} proposed a hierarchical generation model that consists of two stages. In one stage it generates a sequence of words and constituents given a lexicalized syntactic context, and in the other stage it expands the context with the generated sequence. The model repeats these stages iteratively until no constituent is left in the syntactic context.

\subsection{Other Syntax-Related Models}
Though we focus on incorporating syntactic information in the target side, there are several attempts to model source syntax in NMT.
\newcite{t2s} used tree LSTMs to encode source Head-driven Phrase Structure Grammar trees.
\newcite{mss} used word-RNN and structure-RNN to encode source syntax.
\newcite{saed} used an encoder based on top-down and bottom-up tree GRUs, along with a tree-based coverage model.

Another way to utilize syntactic information in NMT is through multi-task learning.
\newcite{l2pt} jointly learn an S2S model and recurrent neural network grammars (RNNGs) on the target side during training.
\newcite{scheduled} proposed a multi-task learning model of translation and parsing.

Several models that generate syntactic structures are proposed in the fields of parsing and language modeling.
\newcite{rnng} introduced RNNGs, which generate sentences with explicit phrase structures.
\newcite{tdlstm} proposed top-down tree LSTMs that generate a dependency tree in top-down, breadth-first order.
\newcite{exact_marginal} proposed a generative shift-reduce dependency parser.

Last but not least, there are a few studies that explore Transformer-based syntactic language models. 
\newcite{sartran-etal-2022-transformer} introduced Transformer Grammars (TGs), which generates linearized constituency trees with recursive syntactic compositions based on Transformer-XL \cite{tfxl}.
\newcite{zhao-etal-2024-dependency} extended TGs and proposed Dependency Transformer Grammars (DTGs), which models a transition sequence of transition-based dependency parsers.

\section{S2S Models with Attention}
Attentional S2S models are composed of an encoder and a decoder, bridged by an attention mechanism.
First, the encoder converts the input sequence into a set of vector representations.
Then the decoder generates the target sequence one by one in a left-to-right order, conditioned on the source representations through the attention mechanism.
While numerous architectures have been proposed to implement such models, we focus on two major variants of such models, namely RNN-based models and the Transformer model.
\subsection{RNN-based Models}
One major approach to attentional encoder-decoder models are based on RNNs \cite{bcb15, luong}.
A typical RNN-based NMT model consists of an encoder based on stacked bidirectional RNNs and a decoder based on stacked unidirectional RNNs coupled with a source attention mechanism. 
The encoder can be formulated as
\begin{align}
\mathbf{s} &= {\rm BiRNN}(\mathbf{E}_{s}[\mathbf{x}]),
\end{align}
where $\mathbf{x}$ is the source sequence, $\mathbf{s}$ is hidden representations of the source sequence and $\mathbf{E}_{s}$ is the source embedding matrix.
The decoder at the $j$-th time step can be formulated as
\begin{align}
h_{j} &= {\rm RNN}(\mathbf{E}_{t}[y_{j-1}], h_{j-1}) \\
c_{j} &= {\rm ATT}(h_{j}, \mathbf{s}) \\
y_{j} &= argmax(g(h_{j}, c_{j})),
\end{align}
where $h$, $c$ and $y$ are a decoder hidden states, a source context and a target word's index, $\mathbf{E}_{t}$ is the target embedding matrix and $g$ is a nonlinear function that outputs the probability of $y_{t}$.
Generally, either LSTMs \cite{lstm} or GRUs \cite{gru} are used as RNN units.

\subsection{Transformer}
More recently, an S2S model called the Transformer is proposed \cite{transformer}.
It has no recurrent components and depends solely on self-attention to handle positional information and history.
Its encoder has multiple layers that consist of a multi-head self-attention sub-layer and a feed-forward sub-layer, and its decoder has a multi-head source-attention sub-layer in addition to the two sub-layers in the encoder.
Both encoder and decoder have a residual connection and layer normalization around each sub-layer.
Since sub-layers in the Transformer contain no positional information, it adds positional encodings to source and target embeddings.
Fixed sinusoidal encodings are generally used.

\section{Proposed Method}
We propose a syntactic NMT model with a decoder that generates a target dependency tree and word sequence.
Unlike previous RNN-based syntactic decoder models that had separate RNN for words and syntactic structures \cite{msra-s2dep, trdec}, our decoder has a single stacked LSTM that handles both dependency structures and surface words, just like S2S models augmented with target syntax. However, it also maintains a stack of actions it should take to complete the tree generation. 
This action stack is used to guide the translation and to inform the decoder with action embeddings.

\subsection{Transition model for top-down incremental decoder}
\begin{figure}[t]
    \centering
    \includegraphics[width=\linewidth]{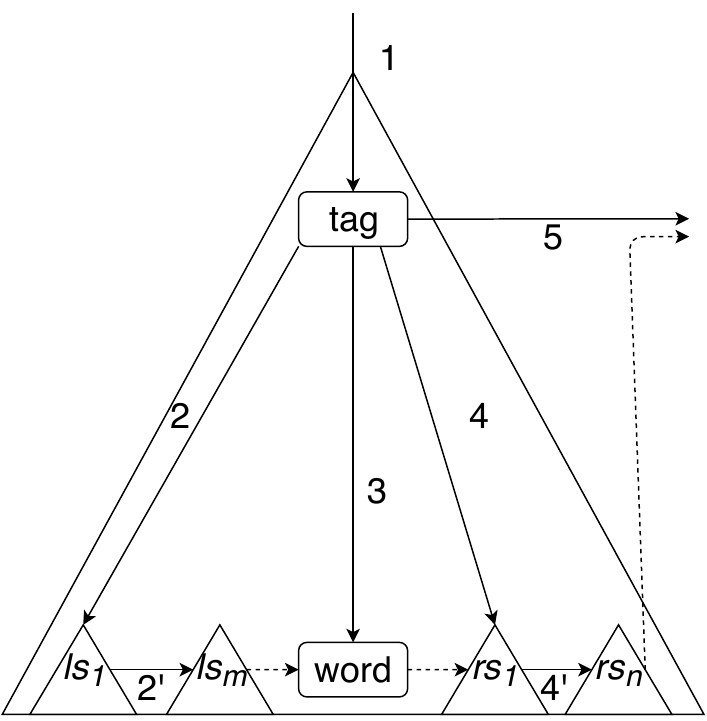}
    \caption{Basic unit of our model that generates a node and a subtree that belongs to it. Each triangle in the figure represents a subtree.}
    \label{fig:basic_unit}
\end{figure}

Our model generates a target dependency tree and words in a top-down, left-to-right order in a recursive manner. 
Figure \ref{fig:basic_unit} shows the basic unit of recursion.
\begin{enumerate}
\item The PoS tag of the head node (\textit{h}) is predicted.
\item If \textit{h} is non-empty, its leftmost child (\textit{lc}) is predicted.
\item If \textit{lc} is non-empty, all left children of \textit{h} and their subtrees (\textit{$ls_{1...m}$}) are generated.
\item If \textit{lc} is empty or all left subtrees are generated, the subword sequence of \textit{h} is generated.
\item After generating \textit{h}'s subword sequence, its most adjacent right child (\textit{rc}) is predicted.
\item If \textit{rc} is non-empty, all right children of \textit{h} and their subtrees (\textit{$rs_{1...n}$}) are generated.
\item If \textit{rc} is empty or all right subtrees are generated, \textit{h}'s sibling is predicted. If \textit{h} has no sibling or all siblings and their subtrees are generated, the procedure returns to \textit{h}'s parent.
\end{enumerate}
The target dependency tree is obtained by recursively applying this procedure. 
Step 7 is omitted for the root node because we assume dependency trees are single-headed.
Note that this model can only generate projective trees. 

\begin{figure}[t]
  \centering
  \includegraphics[width=\linewidth]{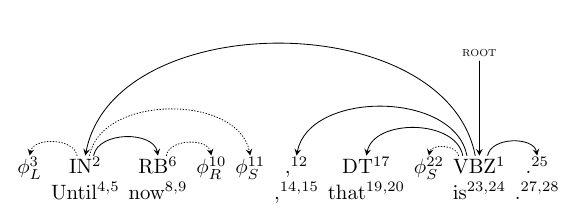}
  \caption{An example dependency tree generated by the procedure in Fig.~\ref{fig:seq}. Most empty symbols are omitted for brevity.}
  \label{fig:tree}
\end{figure}

Figure \ref{fig:tree} shows an example dependency tree, generated by recursively applying the unit depicted in Figure \ref{fig:basic_unit}.
Figure \ref{fig:seq} shows part of the sequence that generates the tree in Figure \ref{fig:tree}. 
Our decoder has four types of actions, LEFT, RIGHT, SIBLING and WORD.
$\phi_{X}$ represents an empty symbol for the action X.
The decoder maintains a stack of actions that are necessary to finish the generation process. 
At each step, the decoder takes the concatenation of action embedding and input embedding (including subwords, PoS tags and empty symbols) and generates an output symbol.
The process starts with a special ROOT embedding as input and with stack of LEFT, WORD and RIGHT (1 in Figure~\ref{fig:seq}).
Basically, the top item of the stack is popped when an empty symbol is generated (3, 5, and 7\textasciitilde 10). 
When the decoder is taking WORD action and a non-empty subword is generated, no change is made to the stack (4 and 11).
If the decoder is taking one of other three actions and a non-empty symbol is generated, the top item of stack is popped, and LEFT, WORD, RIGHT and SIBLING are pushed to the stack in reverse order (2 and 6). 

\begin{figure}[ht]
\small
\centering
\begin{tabular}{l|l|l|l}
 & input & output & stack\\
\hline
1 & ROOT & VBZ & $L|W|R$\\ 
2 & LEFT::VBZ & IN & $L|W|R|S|W|R$\\ 
3 & LEFT::IN & $\phi_{L}$ & $W|R|S|W|R$\\ 
4 & WORD::$\phi_{L}$ & Until & $W|R|S|W|R$\\ 
5 & WORD::Until & $\phi_{W}$ & $R|S|W|R$\\ 
6 & RIGHT::$\phi_{W}$ & RB & $L|W|R|S|S|W|R$\\ 
\multicolumn{3}{c}{...} \\
7 & RIGHT::$\phi_{W}$ & $\phi_{R}$ & $S|S|W|R$\\ 
8 & SIBLING::$\phi_{R}$ & $\phi_{S}$ & $S|W|R$\\
9 & SIBLING::$\phi_{R}$ & , & $L|W|R|S|W|R$\\
\multicolumn{3}{c}{...} \\
10 & SIBLING::$\phi_{R}$ & $\phi_{S}$ & $W|R$\\
11 & WORD::$\phi_{S}$ & is & $W|R$\\
\multicolumn{3}{c}{...} \\
\end{tabular}
\caption{An example action sequence that generates the tree in Fig.~\ref{fig:tree}. 
``::'' stands for a concatenation operation.
$L, W, S$ and $R$ stand for LEFT, WORD, SIBLING and RIGHT actions, respectively.}
\label{fig:seq}
\end{figure}

\subsection{Network Architecture}
Our encoder uses standard bidirectional LSTMs.
We concatenate the outputs of forward and backward LSTMs and pass it as the input for the next layer. 
Linear projection is applied to the output of final layer.

\begin{figure}[t]
\centering
  \includegraphics[width=\linewidth]{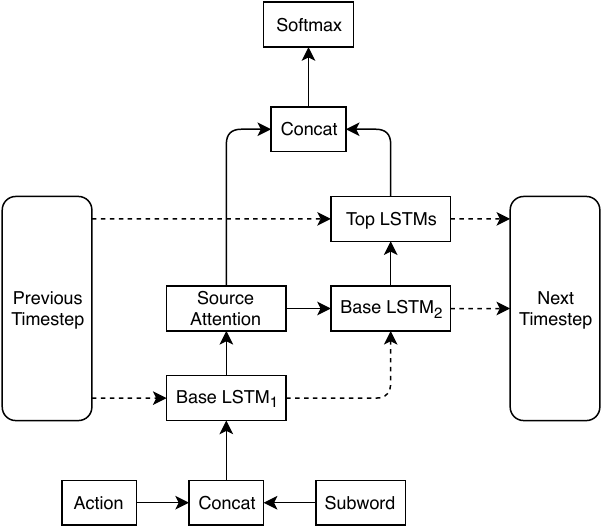}
  \caption{Illustration of the proposed decoder. Dotted lines represent hidden states.}
  \label{fig:decoder}
\end{figure}

Our decoder is based on a variant of LSTM inspired by conditional GRUs \footnote{\url{https://github.com/nyu-dl/dl4mt-tutorial/blob/master/docs/cgru.pdf}} and uses multi-head additive attention \cite{bcb15, bestofboth}.
The calculation of the decoder at the $j$-th timestep can be formulated as
\begin{align}
i_{j} &= concat(\mathbf{E}_{t}[y_{j-1}], \mathbf{A}[a_{j}]) \\
h_{j}^{1'} &= {\rm LSTM}_{1'}(i_{j}, h_{j-1}^{1}) \\
c_{j} &= {\rm ATT}(h_{t}^{1'}, \mathbf{s}) \\
h_{j}^{1} &= {\rm LSTM}_{1}(c_{j}, h_{j}^{1'}) \\
h_{j}^{2...l} &= {\rm LSTM}_{2...l}(h_{j}^{1}, h_{j-1}^{2...l}) \\
y_{j} &= argmax(g(h_{j}^{l}, c_{j})),
\end{align}
where $i$ is the input vector, $a$ is the action index and $\mathbf{A}$ is the action embedding matrix. $h_{j}^{k}$ stands for the hidden states of the $k$-th layer at the $j$-th timestep. 
LSTM cell states are unified with hidden states for brevity.
The overall structure is shown in Figure \ref{fig:decoder}.

\subsection{Loss Function}
Our decoder should generate either subwords or an empty symbol $\phi_{W}$ when it takes the WORD action.
Otherwise, it should generate PoS tags or empty symbols that correspond to other three actions.
In theory, we can use different weight matrix for each action and calculate the loss separately during training. 
However, we choose not to take this option because it complicates the implementation.

\subsection{Difference with Related Models}
Unlike S2S models with target syntax information, our model explicitly maintains a stack that keeps a sequence of actions it should take to finish the translation, and tells the decoder which action it should take through an action embedding.
Hopefully, the information from action embeddings can help the decoder to summarize the partial tree structure that is already built, and the model to generalize better to long inputs compared to purely S2S models.

Unlike models with syntactic decoders that have separate RNNs for tree structure and word sequence, our decoder have one stacked LSTM.
This may enable better information flow between syntactic structures and word sequences compared to models that have separate networks for them.
\newcite{msra-s2dep} predicts dependency structure in a bottom-up order, but our model is top-down.
\newcite{trdec}'s rule-RNN outputs production rules and it cannot generate tree structures that require rules unseen in the training data, whereas our model can generate any projective trees.

\begin{table}[t]
    \centering
    \begin{tabular}{|c|c|c|}
      \hline
      & Topdown & Transformer \\ \hline
      dev & 35.78 & 35.76 \\ \hline
      test & 34.17 & 33.93 \\ \hline
    \end{tabular}
    \caption{Results on WMT16 Ro-En dataset.}
    \label{tab:wmt}
\end{table}

\begin{table}
\begin{tabular}{|c|c|c|}
\hline
Num. verbs & Topdown & Transformer \\ \hline
0 & 39.58 $\pm$ 0.49 & 40.00 $\pm$ 1.48 \\ \hline
1 & 39.42 $\pm$ 1.02 & 38.56 $\pm$ 0.66 \\ \hline
2 & 38.76 $\pm$ 0.64 & 38.62 $\pm$ 0.59 \\ \hline
3 & 37.62 $\pm$ 0.78 & 37.39 $\pm$ 0.73 \\ \hline
4 & 37.15 $\pm$ 0.57 & 37.11 $\pm$ 0.72 \\ \hline
5 & 36.16 $\pm$ 0.79 & 35.94 $\pm$ 0.95 \\ \hline
6 & 35.83 $\pm$ 0.28 & 35.71 $\pm$ 0.23 \\ \hline
7 & 35.94 $\pm$ 0.75 & 34.95 $\pm$ 0.44 \\ \hline
\end{tabular}
\caption{Results on Tilde-subset based on number of verbs in the source sentence, averaged over five test sets with different random seeds.}
\label{tab:tilde}
\end{table}

\section{Experiments}
\label{ssec:layout}
To compare the generalization capabilities of the proposed model and existing S2S models, we train them on three training set with different length thresholds and evaluate them on a dataset that contains sentences with length that are scarce in standard test sets.

\subsection{Datasets}
We trained our models on the WMT 2016 Romanian-to-English dataset \cite{wmt16}. 
We learned the byte-pair encoding (BPE) model \cite{bpe} with 32K merge operations on the concatenation of source and target training data.

To examine how different translation models perform on complex sentences, we prepared a new test data.
In addition to development and test sets of WMT 2016, we evaluated our model also on a subset of the Tilde MODEL corpus \cite{tilde}. 
We binned the sentence pairs in the Tilde corpus based on the number of verbs in the source sentence and randomly extracted up to 500 sentences from each bin.
We call this subset ``Tilde-subset'' hereafter.

As preprocessing, we normalized punctuation and removed Romanian diacritics from the data and tokenized the normalized data. 
We used the Stanford CoreNLP toolkit \cite{corenlp} for English tokenization, and scripts shipped with the Moses decoder\footnote{https://github.com/moses-smt/mosesdecoder} and  wmt16-scripts\footnote{https://github.com/rsennrich/wmt16-scripts} for other preprocessing. 

To obtain dependency trees of the English data, we parsed them using the Berkeley Neural Parser \cite{benepar} and converted the resulting constituency trees to Stanford Dependencies \cite{de2008stanford}.

\subsection{Setup}
We implemented the proposed topdown model and the baseline transformer model using PyTorch \cite{pytorch}.
In the proposed model, word embeddings and LSTMs had 512 units and action embeddings had 32 units. 
The encoder had 3 layers and the decoder had 4 layers ($l$ = 4 in Eq.~10).
We used multi-head attention with 4 heads.
We chose the model with the best BLEU score on newsdev-2016.
No length normalization was used for the topdown model because it affected the BLEU score negatively in preliminary experiments.

We applied weight tying \cite{tying1, tying2} to the source embedding and the input and output embedding of the decoder.

It is reported that part of recent advances in NMT can be attributed to architecture-independent modeling techniques \cite{bestofboth, howmuch}. 
Our decision choices were as follow.
\begin{itemize}
\item We applied Bayesian dropout \cite{dropout} to each LSTM cells in both encoder and decoder with a dropout rate of 0.1.
\item We applied per-gate layer normalization \cite{layernorm} to each LSTM cell. 
\item We applied label smoothing \cite{44903} with uncertainty of 0.1.
\item We used the Adam optimizer \cite{adam} with $\beta_1 = 0.9$,  $\beta_2 = 0.999$, $\epsilon = 10^{-8}$ and varied the learning rate following \newcite{transformer}.
\item We used a mini-batch of 100 examples with delayed update \cite{multi}.
\end{itemize}

The baseline transformer's encoder and decoder both had 6 layers. 

\subsection{Results}
We compute the BLEU score based on tokenized outputs.
Table~\ref{tab:wmt} shows the results on WMT16 development and test sets. 

Table~\ref{tab:tilde} shows the results on Tilde-subset with different number of verbs in the source sentence.
We can see that Transformer performs better when the source sentence has no verb and the proposed model performs better when the source has one or more verbs. 

\section{Discussion}
\label{sec:discussion}
The proposed top-down model lags behind the Transformer in translating sentences with no verb, but it performs better when the input has one or more verbs.

It remains unclear from the result whether the performance drop of the Transformer can be attributed to either of its encoder or decoder, or both are responsible for it. 
According to \newcite{bestofboth}, Transformer encoders perform better than their RNN-based counterparts, even when combined with RNN-based decoders.
It may be interesting to plug the Transformer encoder to the proposed top-down decoder and see if it improves the translation quality as it does in S2S models.

\section{Conclusions and Future Work}
In this paper, we show that despite the recent improvements in NMT, translation of long sentences is still difficult for conventional S2S models especially when the input is substantially longer than sentences in the training data.
This problem has been largely overlooked because the test data in the standard setting does not contain enough amounts of such long sentences to capture it.

To improve the model's generalization capability to model translation of long sentences, we propose a novel syntactic decoder that generates target dependency trees and word sequences in a top-down, left-to-right order.
Our experiments with three different length thresholds for training and a non-standard test set show that the proposed top-down decoder generalizes much better than S2S models to inputs that are substantially longer than those observed during training.

There are several directions of future studies for the proposed model.
First, as mentioned in Chapter \ref{sec:discussion}, it would be interesting to combine the proposed top-down decoder with encoders based on the Transformer model, or more generally the self-attention mechanism.
Second, while our model maintains a stack of action sequences and is informed by action embeddings, its inputs and outputs are largely sequential. 
Introducing some kind of skip connections to reflect partial tree structures may lead to further improvements in its generalization capability.
Third, it is reported that an explicit composition operation is crucial to model long-distance dependencies in the field of language modeling \cite{rnng, kuncoro-etal-2018-lstms}. It would be interesting to introduce the notion of composition to our top-down decoder.
Last but not least, though we use the Stanford dependency converted from automatic parses obtained with the Berkeley parser in this paper, there are many other dependency representations and available parsers. 
It would be interesting to examine to what extent our model depends on the type of dependency representations or the accuracy of automatic parses.
\bibliography{custom}

\end{document}